\documentclass{article} 
\usepackage{nips12submit_e,times}
\usepackage{graphicx,graphics }
\usepackage{amsmath,amsfonts}
\usepackage[numbers,square, sort&compress]{natbib}
\usepackage{epstopdf}
\usepackage{caption}
\usepackage{subcaption}
\usepackage{algorithm}
\usepackage{algpseudocode}

\DeclareMathOperator*{\argmin}{arg\,min}
\DeclareMathOperator*{\argmax}{arg\,max}

\title{Deep Predictive Coding Networks}
\author{
Rakesh Chalasani \hspace*{1cm} Jose C. Principe \\
Department of Electrical and Computer Engineering\\
University of Florida, Gainesville, FL 32611 \\
\texttt{rakeshch@ufl.edu, principe@cnel.ufl.edu}
}

\nipsfinalcopy 

\begin{document}

\maketitle

\begin{abstract}
The quality of data representation in deep learning methods is directly related to the prior model imposed on the representations; however, generally used fixed priors are not capable of adjusting to the context in the data. To address this issue, we propose deep predictive coding networks, a hierarchical generative model that empirically alters priors on the latent representations in a dynamic and context-sensitive manner.  This model captures the temporal dependencies in time-varying signals and uses top-down information to modulate the representation in lower layers. The centerpiece of our model is a novel procedure to infer sparse states of a dynamic network which is used for feature extraction. We also extend this feature extraction block to introduce a pooling function that captures locally invariant representations. When applied on a natural video data, we show that our method is able to learn high-level visual features. We also demonstrate the role of the top-down connections by showing the robustness of the proposed model to structured noise.  
\end{abstract}

\section{Introduction}
\label{sec:intro}
The performance of machine learning algorithms is dependent on how the data is represented. In most methods, the quality of a data representation is itself dependent on prior knowledge imposed on the representation. Such prior knowledge can be imposed using domain specific information, as in SIFT \citep{lowe1999}, HOG \citep{dalal2005}, etc., or in learning representations using fixed priors like sparsity \citep{olshausenfield1996}, temporal coherence \citep{wiskott2002slow}, etc. The use of fixed priors became particularly popular while training deep networks \citep{leeetal2009, kavukcuoglu2010conv, zeiler2010deconvolutional, vincent2010stacked}. In spite of the success of these general purpose priors, they are not capable of adjusting to the context in the data. On the other hand, there are several advantages to having a model that can ``actively" adapt to the context in the data. One way of achieving this is to \emph{empirically alter the priors} in a dynamic and context-sensitive manner. This will be the main focus of this work, with emphasis on visual perception.

Here we propose a predictive coding framework, where a \emph{deep locally-connected} generative model uses ``top-down" information to empirically alter the priors used in the lower layers to perform ``bottom-up" inference. The centerpiece of the proposed model is extracting sparse features from time-varying observations using a \emph{linear dynamical model}. To this end, we propose a novel procedure to infer  \emph{sparse states} (or features) of a dynamical system. We then extend this feature extraction block to introduce a pooling strategy to learn invariant feature representations from the data. In line with other ``deep learning" methods, we  use these basic building blocks to construct a hierarchical model using greedy layer-wise unsupervised learning. The hierarchical model is built such that the output from one layer acts as an input to the layer above. In other words, the layers are arranged in a Markov chain such that the states at any layer are only dependent on the representations in the layer below and above, and are independent of the rest of the model. The overall goal of the dynamical system at any layer is to make the best \emph{prediction} of the representation in the layer below using the top-down information from the layers above and the temporal information from the previous states. Hence, the name \emph{deep predictive coding networks} (DPCN). 

\subsection{Related Work}
The DPCN proposed here is closely related to  models proposed in \citep{raoballard1997, friston2008}, where predictive coding is used as a statistical model to explain cortical functions in the mammalian brain. Similar to the proposed model, they construct hierarchical generative models that seek to infer the underlying causes of the sensory inputs. While \citet{raoballard1997} use an update rule similar to Kalman filter for inference, \citet{friston2008} proposed a general framework considering all the higher-order moments in a continuous time dynamic model. However, neither of the models is capable of extracting discriminative information, namely a sparse and invariant representation, from an image sequence that is helpful for high-level tasks like object recognition. Unlike these models, here we propose an efficient inference procedure to extract locally invariant representation from image sequences and progressively extract more abstract information at higher levels in the model.

Other methods used for building deep models, like restricted Boltzmann machine (RBM) \citep{hinton2006}, auto-encoders \citep{bengio2006, vincent2010stacked} and predictive sparse decomposition \citep{kavukcuoglu2010fast}, are also related to the model proposed here. All these models are constructed on similar underlying principles: (1) like ours, they also use greedy layer-wise unsupervised learning to construct a hierarchical model and (2) each layer consists of an encoder and a decoder. The key to these models is to learn both encoding and decoding concurrently (with some regularization like sparsity \citep{kavukcuoglu2010fast}, denoising \citep{vincent2010stacked} or weight sharing \citep{hinton2006}), while building the deep network as a feed forward model using only the encoder. The idea is to approximate the latent representation using only the feed-forward encoder, while avoiding the decoder which typically requires a more expensive inference procedure. However in DPCN there is no encoder. Instead, DPCN relies on an efficient inference procedure to get a more accurate latent representation. As we will show below, the use of reciprocal top-down and bottom-up connections make the proposed model more robust to structured noise during recognition and also allows it to perform low-level tasks like image denoising. 

To scale to large images, several convolutional models are also proposed in a similar deep learning paradigm \citep{leeetal2009,zeiler2010deconvolutional,kavukcuoglu2010conv}. Inference in these models is applied over an entire image, rather than small parts of the input. DPCN can also be extended to form a convolutional network, but this will not be discussed here. 

\section{Model}
In this section, we begin with a brief description of the general predictive coding framework and proceed to discuss the details of the architecture used in this work. 
The basic block of the proposed model that is pervasive across all layers is a generalized state-space model of the form:
\begin{eqnarray}
		\mathbf{\tilde{y}}_t & = & \mathcal{F}(\mathbf{x}_t) + \mathbf{n}_t \nonumber \\
	\mathbf{x}_{t} & = & \mathcal{G}(\mathbf{x}_{t-1},\mathbf{u}_t) + \mathbf{v}_t 
	\label{eq:generalized_statespace}
\end{eqnarray}
where $\mathbf{\tilde{y}}_t$ is the data and $\mathcal{F}$ and $\mathcal{G}$ are some functions that can be parameterized, say by $\boldsymbol{\theta}$. The terms $\mathbf{u}_t$ are called the \emph{unknown causes}. Since we are usually interested in obtaining abstract information from the observations, the causes are encouraged to have a non-linear relationship with the observations. The hidden states, $\mathbf{x}_t$, then ``mediate the influence of the cause on the output and endow the system with memory" \citep{friston2008}. The terms  $\mathbf{v}_t $ and $\mathbf{n}_t$ are stochastic and model uncertainty. Several such state-space models can now be stacked, with the output from one acting as an input to the layer above, to form a hierarchy. Such an $L$-layered hierarchical model at any time '$t$' can be described as\footnote{When $l = 1$, i.e., at the bottom layer, $\mathbf{u}_t^{(i-1)} = \mathbf{y}_t$, where $\mathbf{y}_t$ the input data.}:
\begin{eqnarray}
	 	\mathbf{u}_t^{(\mathit{l -1})} & = & \mathcal{F}(\mathbf{x}_t^{(\mathit{l})}) + \mathbf{n}_t^{(\mathit{l})}  \qquad \qquad \forall l \in \{1,2,...,L\} \nonumber \\
	\mathbf{x}_{t}^{(\mathit{l})} & = & \mathcal{G}(\mathbf{x}_{t-1}^{(\mathit{l})},\mathbf{u}_t^{(\mathit{l})}) + \mathbf{v}_t^{(\mathit{l})}  \label{eq:generalized_hierarchy}
\end{eqnarray}
The terms $\mathbf{v}_t^{(\mathit{l})}$ and $ \mathbf{n}_t^{(\mathit{l})} $ form stochastic fluctuations at the higher layers and enter each layer independently. In other words, this model forms a Markov chain across the layers, simplifying the inference procedure. Notice how the causes at the lower layer form the ``observations'' to the layer above --- the causes form the link between the layers, and the states link the dynamics over time. The important point in this design is that the higher-level predictions influence the lower levels' inference. The predictions from a higher layer non-linearly enter into the state space model by empirically altering the prior on the causes. In summary, the top-down connections and the temporal dependencies in the state space influence the latent representation at any layer.  

In the following sections, we will first describe a basic computational network, as in (\ref{eq:generalized_statespace}) with a particular form of the functions $\mathcal{F}$ and $\mathcal{G}$. Specifically, we will consider a linear dynamical model with sparse states for encoding the inputs and the state transitions, followed by the non-linear pooling function to infer the causes. Next, we will discuss how to stack and learn a hierarchical model using several of these basic networks. Also, we will discuss how to incorporate the top-down information during inference in the hierarchical model. 

\begin{figure}[t]
\centering
	\begin{subfigure}[b]{0.4\columnwidth}
	\centering
		\includegraphics[width = \textwidth, height = 3cm]{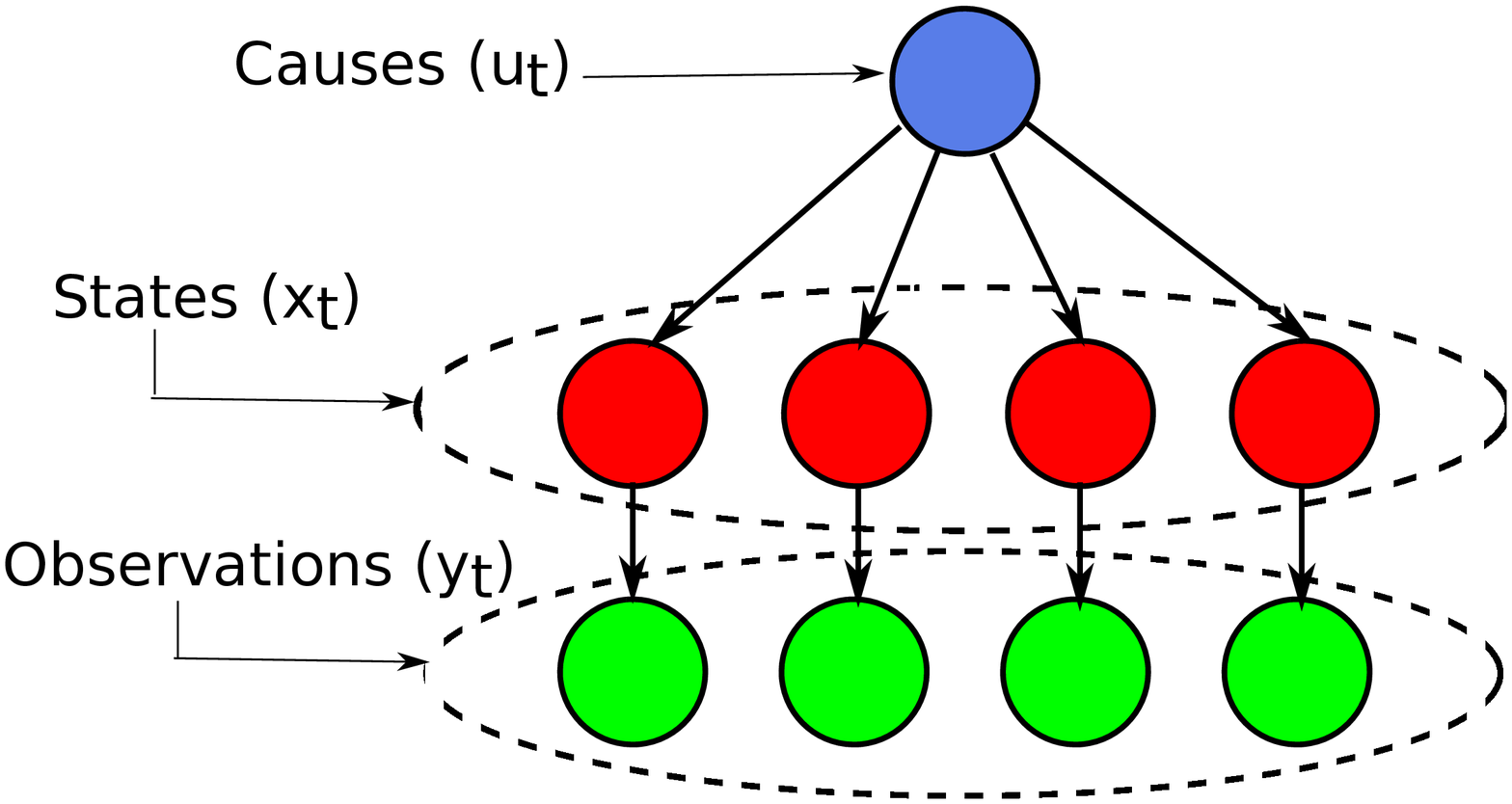}
		\caption{Shows a single layered dynamic network depicting a basic computational block. }
		\label{fig:single_layer}
	\end{subfigure}
	\hspace*{0.5cm}
	\begin{subfigure}[b]{0.5\columnwidth}
	\centering
		\includegraphics[width = \textwidth, height = 4cm]{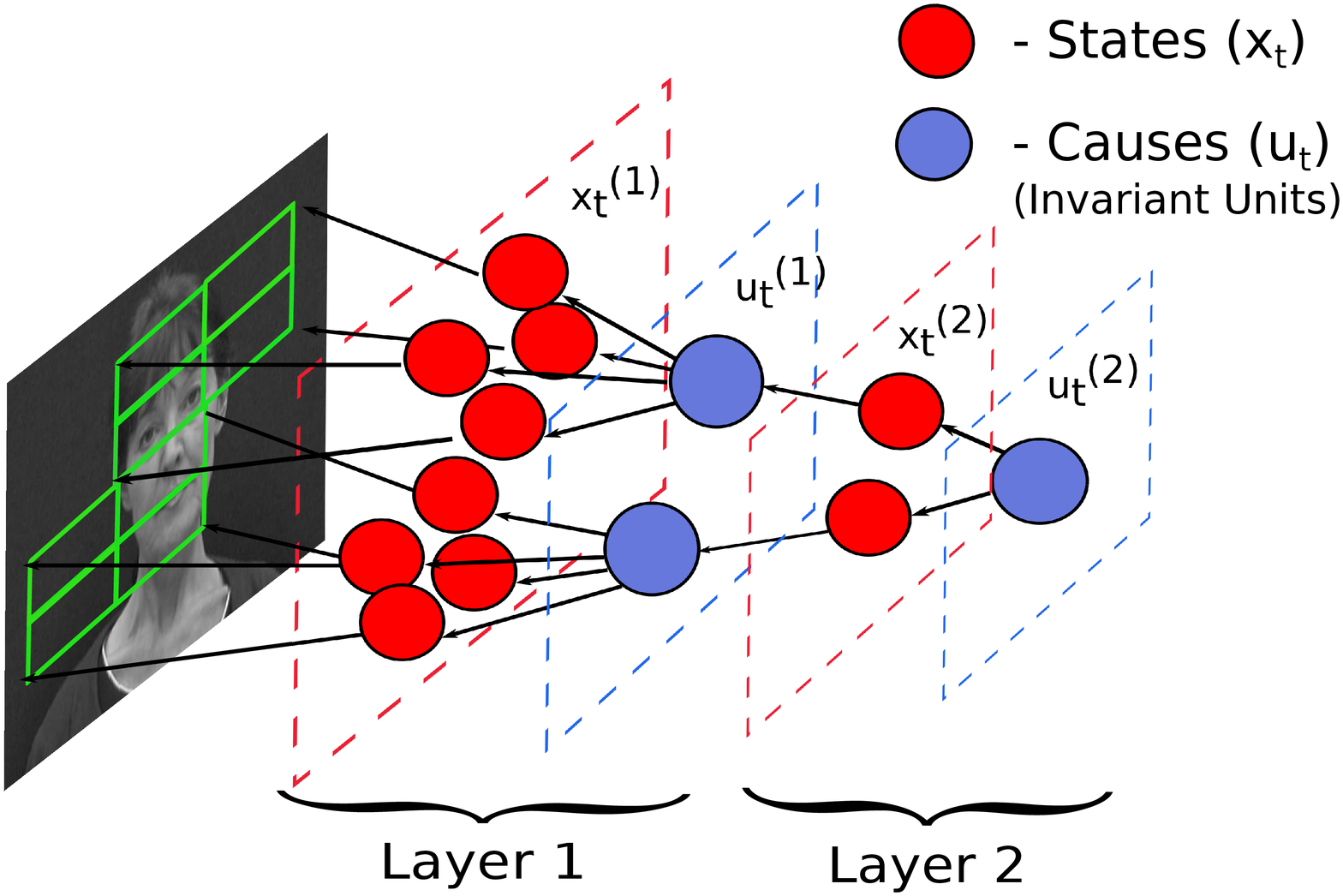}
		\caption{Shows the distributive hierarchical model formed by stacking several basic blocks.}
		\label{fig:architecture}
	\end{subfigure}
	\caption{(a) Shows a single layered network on a group of small overlapping patches of the input video. The green bubbles indicate a group of inputs ($\mathbf{y}_t^{(n)}, \forall n$), red bubbles indicate their corresponding states ($\mathbf{x}_t^{(n)}$) and the blue bubbles indicate the causes ($\mathbf{u}_t$) that pool all the states within the group. (b) Shows a two-layered hierarchical model constructed by stacking several such basic blocks. For visualization no overlapping is shown between the image patches here, but overlapping patches are considered during actual implementation.}
	\label{fig:single_and_hierarchy}
\end{figure}


\subsection{Dynamic network}
\label{sec:energy_function}
To begin with, we consider a dynamic network to extract features from a small part of a video sequence. Let $\{\mathbf{y}_1,\mathbf{y}_2,...,\mathbf{y}_t,...\} \in \mathbb{R}^P$ be a $P$-dimensional sequence of a patch extracted from the same location across all the frames in a video\footnote{Here $\mathbf{y}_t$ is a vectorized form of $\sqrt{P}\times\sqrt{P}$ square patch extracted from a frame at time $t$.} . To process this, our network consists of two distinctive parts (see Figure.\ref{fig:single_layer}): feature extraction (inferring states) and pooling (inferring causes). For the first part, sparse coding is used in conjunction with a \emph{linear state space model} to map the inputs $\mathbf{y}_t$ at time $t$ onto an over-complete dictionary of $K$-filters, $\mathbf{C} \in \mathbb{R}^{P \times K} (K > P)$, to get sparse states $\mathbf{x}_t \in \mathbb{R}^K$. To keep track of the dynamics in the latent states  we use a linear function with state-transition matrix $\mathbf{A} \in \mathbb{R}^{K\times K}$. More formally, inference of the features $\mathbf{x}_t$ is performed by finding a representation that minimizes the energy function:
\begin{eqnarray}
	E_1(\mathbf{x}_t, \mathbf{y}_t, \mathbf{C}, \mathbf{A})& = & \|\mathbf{y}_t - \mathbf{C} \mathbf{x}_t\|_2^2 + \lambda \|\mathbf{x}_t - \mathbf{A}\mathbf{x}_{t-1}\|_1 + \boldsymbol{\gamma} \|\mathbf{x}_t\|_1
	\label{eq:state_cost}
\end{eqnarray}
Notice that the second term involving the state-transition is also constrained to be sparse to make the state-space representation consistent.

Now, to take advantage of the spatial relationships in a local neighborhood, a small group of states $\mathbf{x}_t^{(n)}$, where $n \in \{1,2,...N\}$ represents a set of contiguous patches w.r.t. the position in the image space, are added (or \emph{sum pooled}) together. Such pooling of the states may be lead to local translation invariance. On top this, a $D$-dimensional causes $\mathbf{u}_t \in \mathbb{R}^{D}$ are inferred from the pooled states to obtain representation that is invariant to more complex local transformations like rotation, spatial frequency, etc. In line with \citep{karklin2005}, this invariant function is learned such that it can capture the dependencies between the components in the pooled states. Specifically, the causes $\mathbf{u}_t$ are inferred by minimizing the energy function:
\begin{eqnarray}
	E_2(\mathbf{u}_t,\mathbf{x}_t,\mathbf{B})& = &  \sum_{n=1}^{N} \Big(\sum_{k=1}^K |\gamma_{k}  \cdot {x}_{t,k}^{(n)}|  \Big) + \beta \|\mathbf{u}_{t}\|_1 \\
	 	\gamma_{k}& = & \gamma_0 \Big[ \frac{1 + exp(-[\mathbf{B} \mathbf{u}_{t}]_{k})}{2}\Big] \nonumber 
	 	\label{eq:cause_cost}
\end{eqnarray}
where $\gamma_0 >0$ is some constant. Notice that here $\mathbf{u}_t$ multiplicatively interacts with the accumulated states through $\mathbf{B}$, modeling the shape of the sparse prior on the states. Essentially, the invariant matrix $\mathbf{B}$ is adapted such that each component $\mathbf{u}_t$ connects to a group of components in the accumulated states that co-occur frequently.  In other words, whenever a component in $\mathbf{u}_t$ is active it lowers the coefficient of a set of components in $\mathbf{x}_t^{(n)}, \forall n$, making them \emph{more likely} to be active. Since co-occurring components typically share some common statistical regularity, such activity of $\mathbf{u}_t$ typically leads to locally invariant representation \citep{karklin2005}. 

Though the two cost functions are presented separately above, we can combine both to devise a unified energy function of the form:
\begin{align}
	{E}(\mathbf{x}_t,\mathbf{u}_t,\boldsymbol{\theta}) = & \sum_{n=1}^{N} \Big( \frac{1}{2} \|\mathbf{y}_t^{(n)} - \mathbf{C}\mathbf{x}_t^{(n)}\|_2^2	+ \mathbf{\lambda}  \|\mathbf{x}_t^{(n)} - \mathbf{A}\mathbf{x}_{t-1}^{(n)}\|_1 + \sum_{k=1}^K |\gamma_{t,k} \cdot {x}_{t,k}^{(n)}|  \Big) + \beta \|\mathbf{u}_{t}\|_1 
		\label{eq:single_cost_func} \\
	 	\gamma_{t,k} = &\gamma_0 \Big[ \frac{1 + exp(-[\mathbf{B} \mathbf{u}_{t}]_{k})}{2}\Big] \nonumber
\end{align}
where $\boldsymbol{\theta} = \{\mathbf{A},\mathbf{B},\mathbf{C}\}$. As we will discuss next, both $\mathbf{x}_t$ and $\mathbf{u}_t$ can be inferred concurrently from (\ref{eq:single_cost_func}) by alternatively updating one while keeping the other fixed using an efficient proximal gradient method.  

\subsection{Learning}
\label{sec:learning}
To learn the parameters in (\ref{eq:single_cost_func}), we alternatively minimize ${E}(\mathbf{x}_t,\mathbf{u}_t,\boldsymbol{\theta})$ using a procedure similar to block co-ordinate descent. We first infer the latent variables $(\mathbf{x}_t, \mathbf{u}_t)$ while keeping the parameters fixed and then update the parameters $\boldsymbol{\theta}$ while keeping the variables fixed. This is done until the parameters converge. We now discuss separately the inference procedure and how we update the parameters using a gradient descent method with the fixed variables.



\subsubsection{Inference}
\label{sec:infer}
We jointly infer both $\mathbf{x}_t$ and $\mathbf{u}_t$ from (\ref{eq:single_cost_func}) using proximal gradient methods, taking alternative gradient descent steps to update one while holding the other fixed. In other words, we alternate between updating $\mathbf{x}_t$ and $\mathbf{u}_t$ using a single update step to minimize $E_1$ and $E_2$, respectively. However, updating $\mathbf{x}_t$ is relatively more involved. So, keeping aside the causes, we first focus on inferring sparse states alone from $E_1$, and then go back to discuss the joint inference of both the states and the causes.


\textbf{Inferring States:} Inferring sparse states, given the parameters, from a linear dynamical system forms the crux of our model. This is performed by finding the solution that minimizes the energy function $E_1$ in (\ref{eq:state_cost}) with respect to the states $\mathbf{x}_t$ (while keeping the sparsity parameter $\boldsymbol{\gamma}$ fixed). 
Here there are two priors of the states: the temporal dependence and the sparsity term. Although this energy function $E_1$ is convex in $\mathbf{x}_t$, the presence of two non-smooth terms makes it hard to use standard optimization techniques used for sparse coding alone. A similar problem is solved using dynamic programming \cite{angelo2009}, homotopy \cite{charles2011} and Bayesian sparse coding \cite{sejdinovic2010}; however, the optimization used in these models is computationally expensive for use in large scale problems like object recognition. 

To overcome this, inspired by the method proposed in \citep{chen2012smoothing} for structured sparsity, we propose an approximate solution that is consistent and able to use efficient solvers like fast iterative shrinkage thresholding alogorithm (FISTA) \cite{fista2009}. The key to our approach is to first use Nestrov's smoothness method \citep{nesterov2005smooth, chen2012smoothing} to approximate the non-smooth state transition term. The resulting energy function is a convex and continuously differentiable function in $\mathbf{x}_t$ with a sparsity constraint, and hence, can be efficiently solved using proximal methods like FISTA. 

To begin, let $\Omega(\mathbf{x}_t) = \|\mathbf{e}_t\|_1$ where $\mathbf{e}_t = (\mathbf{x}_t - \mathbf{A}\mathbf{x}_{t-1})$. The idea is to find a smooth approximation to this function $\Omega(\mathbf{x}_t)$ in $\mathbf{e}_t$. Notice that, since $\mathbf{e}_t$ is a linear function on $\mathbf{x}_t$, the approximation will also be smooth w.r.t. $\mathbf{x}_t$. Now, we can re-write  $\Omega(\mathbf{x}_t)$ using the dual norm of $\ell_1$ as
$$ 
\Omega(\mathbf{x}_t)  =  \argmax\limits_{\|\boldsymbol{\alpha}\|_{\infty} \leq 1} \boldsymbol{\alpha}^{T}\mathbf{e}_t 
$$
where $\boldsymbol{\alpha} \in \mathbb{R}^{k}$. Using the smoothing approximation from \citet{nesterov2005smooth} on $\Omega(\mathbf{x}_t)$:
\begin{eqnarray}
	\Omega(\mathbf{x}_t) & \approx & f_{\mu}(\mathbf{e}_t) = \argmax\limits_{\|\boldsymbol{\alpha}\|_{\infty} \leq 1} [\boldsymbol{\alpha}^{T}\mathbf{e}_t - \mu d(\boldsymbol{\alpha})] 
	\label{eq:nestrov_approx}
\end{eqnarray}
where $d(\cdot) = \frac{1}{2} \|\boldsymbol{\alpha}\|_2^2$ is a smoothing function and $\mu$ is a smoothness parameter. From Nestrov's theorem \citep{nesterov2005smooth}, it can be shown that $f_{\mu}(\mathbf{e}_t)$ is convex and continuously differentiable in $\mathbf{e}_t$ and the gradient of $f_{\mu}(\mathbf{e}_t)$ with respect to $\mathbf{e}_t$ takes the form
\begin{eqnarray}
	\nabla_{\mathbf{e}_t} f_{\mu}(\mathbf{e}_t) & = & \boldsymbol{\alpha}^{*} 
\end{eqnarray}
	where $\boldsymbol{\alpha}^{*}$ is the optimal solution to $f_{\mu}(\mathbf{e}_t) = \argmax\limits_{\|\boldsymbol{\alpha}\|_{\infty} \leq 1} [\boldsymbol{\alpha}^{T}\mathbf{e}_t - \mu d(\boldsymbol{\alpha})]$ \footnote{Please refer to the supplementary material for the exact form of $\boldsymbol{\alpha}*$.}. This implies, by using the chain rule, that $f_{\mu}(\mathbf{e}_t)$ is also convex and continuously differentiable in $\mathbf{x}_t$ and with the same gradient. 
	
With this smoothing approximation, the overall cost function from (\ref{eq:state_cost}) can now be re-written as
\begin{eqnarray}
	\mathbf{x}_t = \argmin_{\mathbf{x}_t} \frac{1}{2} \|\mathbf{y}_t - \mathbf{C}\mathbf{x}_t\|_2^2	+ \mathbf{\lambda} f_{\mu}(\mathbf{e}_t) + \gamma \|\mathbf{x}_t\|_1
	\label{eq:state_smooth_cost}
\end{eqnarray}
with the smooth part $h(\mathbf{x}_t) = \frac{1}{2} \|\mathbf{y}_t - \mathbf{C}\mathbf{x}_t\|_2^2	+ \mathbf{\lambda} f_{\mu}(\mathbf{e}_t)$ whose gradient with respect to $\mathbf{x}_t$ is given by
	\begin{eqnarray}
		\nabla_{\mathbf{x}_t} h(\mathbf{x}_t) = \mathbf{C}^{T}(\mathbf{y}_t - \mathbf{C}\mathbf{x}_t) + \lambda \boldsymbol{\alpha}^{*} 
		\label{eq:state_grad}
\end{eqnarray}	 
Using the gradient information in (\ref{eq:state_grad}), we solve for $\mathbf{x}_t$ from (\ref{eq:state_smooth_cost}) using FISTA \citep{fista2009}.

\textbf{Inferring Causes:} Given a group of state vectors, $\mathbf{u}_t$ can be inferred by minimizing $E_{2}$, where we define a generative model that modulates the sparsity of the \emph{pooled} state vector, $\sum_n |\mathbf{x}^{(n)}|$. Here we observe that FISTA can be readily applied to infer $\mathbf{u}_t$, as the smooth part of the function $E_2$:
\begin{eqnarray}
	h(\mathbf{u}_t) & = & \sum_{k=1}^K  \Big(\gamma_0 \Big[ \frac{1 + \exp(-[\mathbf{B} \mathbf{u}_{t}]_{k})}{2}\Big] \cdot \sum_{n=1}^{N} | {x}_{t,k}^{(n)}|  \Big)
	\label{eq:causes_smooth}
\end{eqnarray}
is convex, continuously differentiable and Lipschitz in $\mathbf{u}_t$ \citep{gregor2011} \footnote{The matrix $\mathbf{B}$ is initialized with non-negative entries and continues to be non-negative without any additional constraints \citep{gregor2011}.}.  Following \citep{fista2009}, it is easy to obtain a bound on the convergence rate of the solution. 

\textbf{Joint Inference:} We showed thus far that both $\mathbf{x}_t$ and $\mathbf{u}_t$ can be inferred from their respective energy functions using a first-order proximal method called FISTA. However, for joint inference we have to   minimize the combined energy function in (\ref{eq:single_cost_func}) over both $\mathbf{x}_t$ and $\mathbf{u_t}$.  We do this by alternately updating $\mathbf{x}_t$ and $\mathbf{u}_t$ while holding the other fixed and using a \emph{single} FISTA update step at each iteration. It is important to point out that the internal FISTA step size parameters are maintained between iterations. This procedure is equivalent to alternating minimization using gradient descent. Although this procedure no longer guarantees convergence of both $\mathbf{x}_t$ and $\mathbf{u}_t$ to the optimal solution, in all of our simulations it lead to a reasonably good solution. Please refer to Algorithm.~\ref{algo:fista} (in the supplementary material) for details. Note that, with the alternating update procedure, each $\mathbf{x}_t$ is now influenced by the feed-forward observations, temporal predictions and the feedback connections from the causes. 

\subsubsection{Parameter Updates}
With $\mathbf{x}_t$ and $\mathbf{u}_t$ fixed, we update the parameters by minimizing $E$ in (\ref{eq:single_cost_func}) with respect to $\boldsymbol{\theta}$. Since the inputs here are a time-varying sequence, the parameters are updated using dual estimation filtering \citep{nelson2000}; i.e., we put an additional constraint on the parameters such that they follow a state space equation of the form:
\begin{eqnarray}
	\boldsymbol{\theta}_t & = & \boldsymbol{\theta}_{t-1} + \mathbf{z}_t
\end{eqnarray}
where $\mathbf{z}_t$ is Gaussian transition noise over the parameters. This keeps track of their temporal relationships. Along with this constraint, we update the parameters using gradient descent. Notice that with a fixed $\mathbf{x}_t$ and $\mathbf{u}_t$, each of the parameter matrices can be updated independently. Matrices $\mathbf{C}$ and $\mathbf{B}$ are column normalized after the update to avoid any trivial solution. 

\textbf{Mini-Batch Update:} To get faster convergence, the parameters are updated after performing inference over a large sequence of inputs instead of at every time instance. With this ``batch" of signals, more sophisticated gradient methods, like conjugate gradient, can be used and, hence, can lead to more accurate and faster convergence.

\subsection{Building a hierarchy} 

So far the discussion is focused on encoding a small part of a video frame using a single stage network. To build a hierarchical model, we use this single stage network as a basic building block and arrange them up to form a \emph{tree} structure (see Figure.\ref{fig:architecture}). To learn this hierarchical model, we adopt a greedy layer-wise procedure like many other deep learning methods \citep{hinton2006, kavukcuoglu2010conv, vincent2010stacked}. Specifically, we use the following strategy to learn the hierarchical model.  

For the first (or bottom) layer, we learn a dynamic network as described above over a group of small patches from a video. We then take this learned network and \emph{replicate} it at several places on a larger part of the input frames (similar to weight sharing in a convolutional network \citep{lecun1989}). The outputs (causes) from each of these replicated networks are considered as inputs to the layer above. Similarly, in the second layer the inputs are again grouped together (depending on the spatial proximity in the image space) and are used to train another dynamic network. Similar procedure can be followed to build more higher layers. 

We again emphasis that the model is learned in a layer-wise manner, i.e., there is no top-down information while learning the network parameters. Also note that, because of the \emph{pooling} of the states at each layers, the receptive field of the causes becomes progressively larger with the depth of the model.

\subsection{Inference with top-down information}
\label{sec:top_down}
With the parameters fixed, we now shift our focus to inference in the hierarchical model with the top-down information. As we discussed above, the layers in the hierarchy are arranged in a Markov chain, i.e., the variables at any layer are only influenced by the variables in the layer below and the layer above. Specifically, the states $\mathbf{x}_t^{(l)}$ and the causes $\mathbf{u}_t^{(l)}$ at layer $l$ are inferred from $\mathbf{u}_t^{(l-1)}$ and are influenced by $\mathbf{x}_t^{(l+1)}$ (through the prediction term $\mathbf{C}^{(l+1)}\mathbf{x}_t^{(l+1)}$) \footnote{The suffixes $n$ indicating the group are considered implicit here to simplify the notation.}. Ideally, to perform inference in this hierarchical model, all the states and the causes have to be updated simultaneously depending  on the present state of all the other layers until the model reaches equilibrium \citep{friston2008}. However, such a procedure can be very slow in practice. Instead, we propose an approximate inference procedure that only requires a single top-down flow of information and then a single bottom-up inference using this top-down information.

For this we consider that at any layer $l$ a group of input $\mathbf{u}_t^{(l-1,n)}, \forall n \in \{1,2,...,N\}$ are encoded using a group of states $\mathbf{x}_t^{(l,n)}, \forall n$ and the causes $\mathbf{u}_t^{(l)}$ by minimizing the following energy function:
\begin{eqnarray}
	{E}_{l}(\mathbf{x}_t^{(l)},\mathbf{u}_t^{(l)},\boldsymbol{\theta}^{(l)}) & = & \sum_{n=1}^{N} \Big( \frac{1}{2} \|\mathbf{u}_t^{(l-1,n)} - \mathbf{C}^{(l)}\mathbf{x}_t^{(l,n)}\|_2^2	+ \mathbf{\lambda}  \|\mathbf{x}_t^{(l,n)} - \mathbf{A}^{(l)}\mathbf{x}_{t-1}^{(l,n)}\|_1 \nonumber \\
		& & \qquad \qquad  + \sum_{k=1}^K |\gamma_{t,k}^{(l)} \cdot {x}_{t,k}^{(l,n)}|  \Big) + \beta \|\mathbf{u}_{t}^{(l)}\|_1 + \frac{1}{2}\|\mathbf{u}_{t}^{(l)} - \mathbf{\hat{u}}_t^{(l+1)}\|_2^2
		\label{eq:cost_func} \\
	 	\gamma_{t,k}^{(l)} & = &\gamma_0 \Big[ \frac{1 + exp(-[\mathbf{B}^{(l)} \mathbf{u}_{t}^{(l)}]_{k})}{2}\Big] \nonumber
\end{eqnarray}
where $\boldsymbol{\theta}^{(l)} = \{\mathbf{A}^{(l)},\mathbf{B}^{(l)},\mathbf{C}^{(l)}\}$. Notice the additional term involving $\mathbf{\hat{u}}_t^{(l+1)}$ when compared to (\ref{eq:single_cost_func}). This comes from the top-down information, where we call $\mathbf{\hat{u}}_t^{(l+1)}$ as the top-down prediction of the causes of layer $(l)$ using the previous states in layer $(l+1)$. Specifically, before the ``arrival" of a new observation at time $t$, at each layer $(l)$ (starting from the top-layer) we first propagate the most likely causes to the layer below using the state at the previous time instance $\mathbf{x}_{t-1}^{(l)}$ and the predicted causes $\mathbf{\hat{u}}_t^{(l+1)}$. More formally, the top-down prediction at layer $l$ is obtained as
\begin{eqnarray}
			\mathbf{\hat{u}}_t^{(l)} & = & \mathbf{C}^{(l)}\mathbf{\hat{x}}_t^{(l)} \nonumber \\
	\textrm{where} \quad	\mathbf{\hat{x}}_t^{(l)} & = & \argmin_{\mathbf{x}_t^{(l)}} \lambda^{(l)}\|\mathbf{x}_t^{(l)} - \mathbf{A}^{(l)}\mathbf{x}_{t-1}^{(l)}\|_1 + \gamma_0 \sum_{k=1}^K |{\hat{\gamma}_{t,k}} \cdot \mathbf{x}_{t,k}^{(l)}| \label{eq:predict_cost} \\
		\textrm{and} \quad \hat{\gamma}_{t,k} & = & ( \exp(-[\mathbf{B}^{(l)}\mathbf{\hat{u}}_t^{(l+1)}]_{k}))/2 \nonumber
	\end{eqnarray}	 
	 At the top most layer, $L$, a ``bias" is set such that $\mathbf{\hat{u}}_t^{(L)} = \mathbf{\hat{u}}_{t-1}^{(L)}$, i.e., the top-layer induces some temporal coherence on the final outputs.
	 From (\ref{eq:predict_cost}), it is easy to show that the predicted states for layer ${l}$ can be obtained as
	\begin{eqnarray}
		{\hat{x}}_{t,k}^{(l)} &=& 
		\begin{cases}
			[\mathbf{A}^{(l)}\mathbf{x}_{t-1}^{(l)}]_{k}, & \gamma_0\gamma_{t,k} < \lambda^{(l)} \\
			0, & \gamma_0\gamma_{t,k} \geq \lambda^{(l)}
		\end{cases} 
	\end{eqnarray}
These predicted causes $\mathbf{\hat{u}}_t^{(l)}, \  \forall l \in \{1,2,...,L\}$ are substituted in (\ref{eq:cost_func}) and a single layer-wise bottom-up inference is performed as described in section \ref{sec:infer} \footnote{Note that the additional term $\frac{1}{2}\|\mathbf{u}_{t}^{(l)} - \mathbf{\hat{u}}_t^{(l+1)}\|_2^2$ in the energy function only leads to a minor modification in the inference procedure, namely this has to be added to $h(\mathbf{u}_t)$ in (\ref{eq:causes_smooth}).}. The \emph{combined prior} now imposed on the causes, $\beta \|\mathbf{u}_{t}^{(l)}\|_1 + \frac{1}{2}\|\mathbf{u}_{t}^{(l)} - \mathbf{\hat{u}}_t^{(l+1)}\|_2^2$, is similar to the elastic net prior discussed in  \citep{zou2005regularization}, leading to a smoother and biased estimate of the causes. 
	 
\section{Experiments}

\subsection{Receptive fields of causes in the hierarchical model}
\label{sec:results_recept}
\begin{figure}[t]
\centering
	\begin{subfigure}[b]{0.45\textwidth}
	\centering
		\includegraphics[width = \columnwidth]{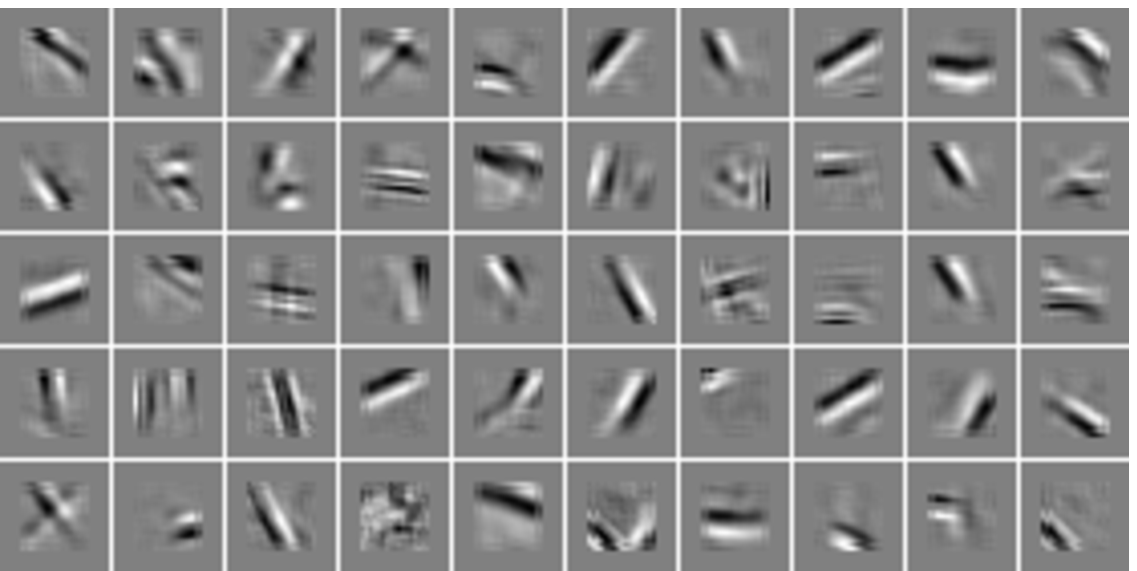}
		\caption{Layer 1 invariant matrix, $\mathbf{B}^{(1)}$}
	\end{subfigure}
	\hspace*{0.5cm}
	\begin{subfigure}[b]{0.45\textwidth}
	\centering
		\includegraphics[width = \columnwidth]{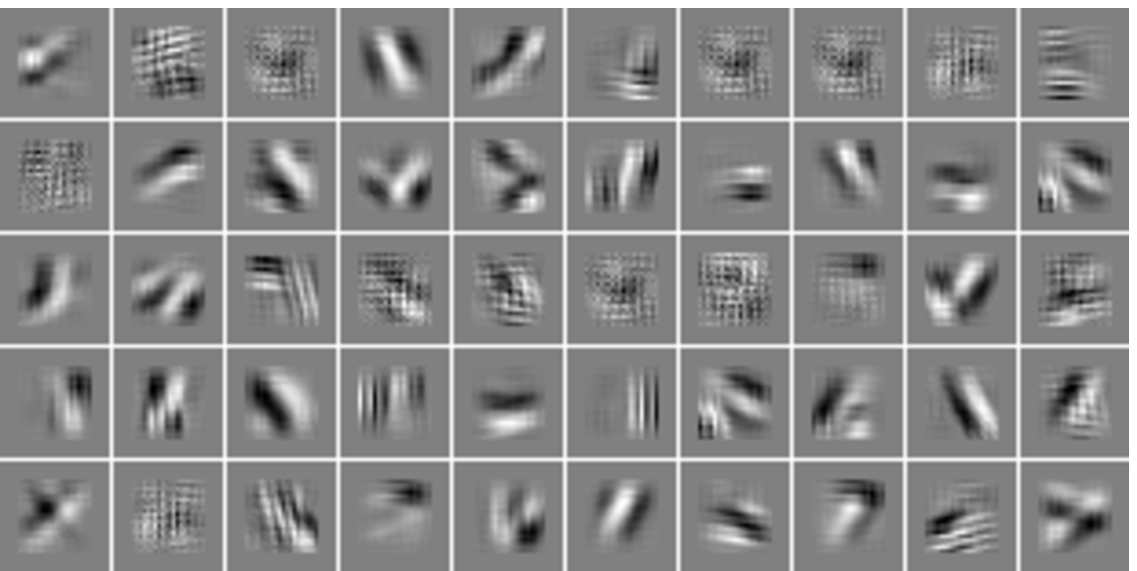}
		\caption{Layer 2 invariant matrix, $\mathbf{B}^{(2)}$}
	\end{subfigure}
	\caption{Visualization of the receptive fields of the invariant units learned in (a) layer 1  and (b) layer 2 when trained on natural videos. The receptive fields are constructed as a weighted combination of the dictionary of filters at the bottom layer.}
	\label{fig:receptive_fields}
\end{figure}
Firstly, we would like to test the ability of the proposed model to learn complex features in the higher-layers of the model. For this we train a two layered network from a natural video. Each frame in the video was first contrast normalized as described in \citep{kavukcuoglu2010fast}. Then, we train the first layer of the model on $4$ overlapping contiguous $15 \times 15$ pixel patches from this video; this layer has 400 dimensional states and 100 dimensional causes. The causes pool the states related to all the $4$ patches. The separation between the overlapping patches here was $2$ pixels, implying that the receptive field of the causes in the first layer is $17 \times 17$ pixels. Similarly, the second layer is trained on $4$ causes from the first layer obtained from $4$ overlapping $17 \times 17$ pixel patches from the video. The separation between the patches here is $3$ pixels, implying that the receptive field of the causes in the second layer is $20 \times 20$ pixels. The second layer contains 200 dimensional states and 50 dimensional causes that pools the states related to all the $4$ patches.  

Figure~\ref{fig:receptive_fields} shows the visualization of the receptive fields of the invariant units (columns of matrix $\mathbf{B}$) at each layer. We observe that each dimension of causes in the first layer represents a group of primitive features (like inclined lines) which are localized in orientation or position \footnote{Please refer to supplementary material for more results.}. Whereas, the causes in the second layer represent more complex features, like corners, angles, etc. These filters are consistent with the previously proposed methods like \citet{leeetal2009} and \citet{zeiler2010deconvolutional}.   

\subsection{Role of top-down information}
\begin{figure}[t]
	\centering
	\begin{subfigure}[b]{0.45\textwidth}
		\centering
		\includegraphics[width = \columnwidth]{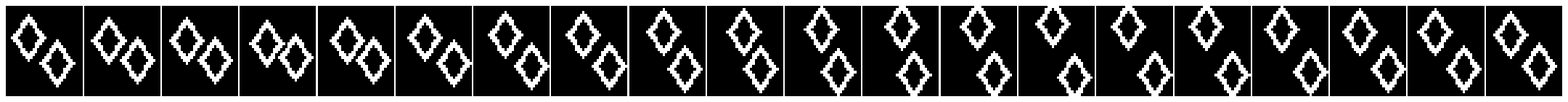} \\
		\includegraphics[width = \columnwidth]{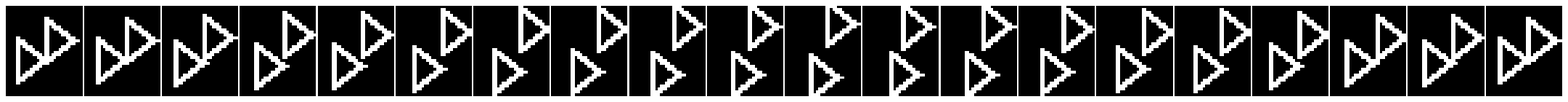} \\
		\includegraphics[width = \columnwidth]{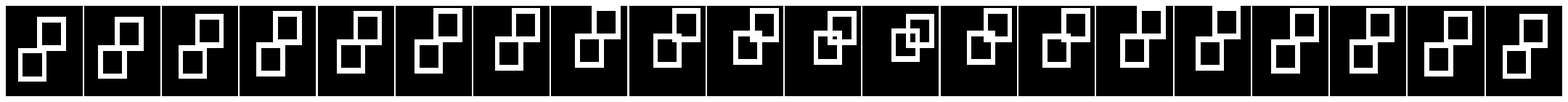}
		\caption{Clear Sequences}
	\end{subfigure}
		\begin{subfigure}[b]{0.45\textwidth}
		\centering
		\includegraphics[width = \columnwidth]{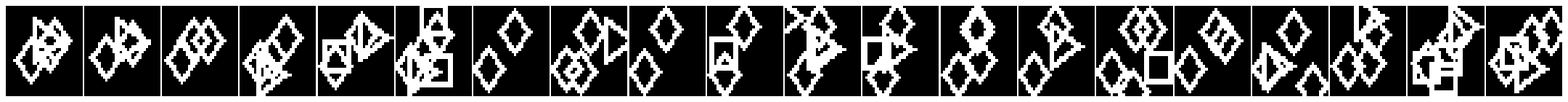} \\
		\includegraphics[width = \columnwidth]{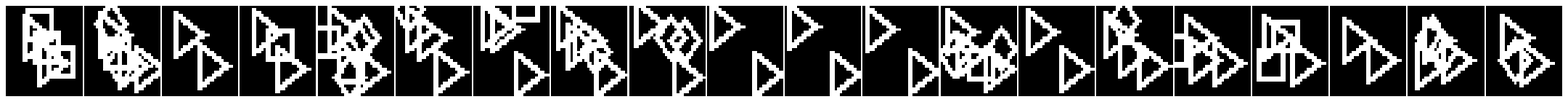} \\
		\includegraphics[width = \columnwidth]{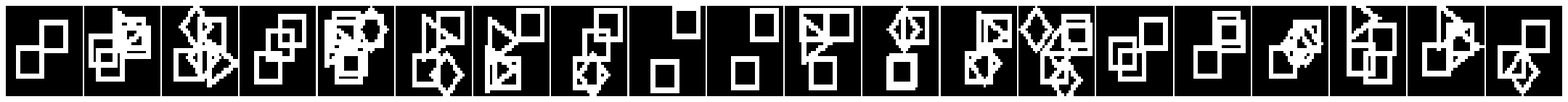}
		\caption{Corrupted Sequences}
		\label{fig:toy_corrupt}
	\end{subfigure}
	\caption{Shows part of the (a) clean and (b) corrupted video sequences constructed using three different shapes. Each row indicates one sequence.}
	\label{fig:objects}
\end{figure}

In this section, we show the role of the top-down information during inference, particularly in the presence of structured noise. Video sequences consisting of objects of three different shapes (Refer to Figure \ref{fig:objects}) were constructed. The objective is to classify each frame as coming from one of the three different classes. For this, several $32 \times 32$ pixel 100 frame long sequences were made using two objects of the same shape bouncing off each other and the ``walls''. Several such sequences were then concatenated to form a 30,000 long sequence. We train a two layer network using this sequence. First, we divided each frame into $12 \times 12$ patches with neighboring patches overlapping by 4 pixels; each frame is divided into 16 patches. The bottom layer was trained such the $12 \times 12$ patches were used as inputs and were encoded using a 100 dimensional state vector. A $4$ contiguous neighboring patches were pooled to infer the causes that have 40 dimensions. The second layer was trained with $4$ first layer causes as inputs, which were itself inferred from $20 \times 20$ contiguous overlapping blocks of the video frames. The states here are 60 dimensional long and the causes have only 3 dimensions. It is important to note here that the receptive field of the second layer causes encompasses the entire frame.    

We test the performance of the DPCN in two conditions. The first case is with 300 frames of clean video, with 100 frames per shape, constructed as described above. We consider this as a single video without considering any discontinuities. In the second case, we corrupt the clean video with ``structured" noise, where we randomly pick a number of objects from same three shapes with a Poisson distribution (with mean 1.5) and add them to each frame independently at a random locations. There is no correlation between any two consecutive frames regarding where the ``noisy objects" are added (see Figure.\ref{fig:toy_corrupt}). 

\begin{figure}[t]
\centering
	\begin{subfigure}[b]{0.3\columnwidth}
		\includegraphics[width = \columnwidth]{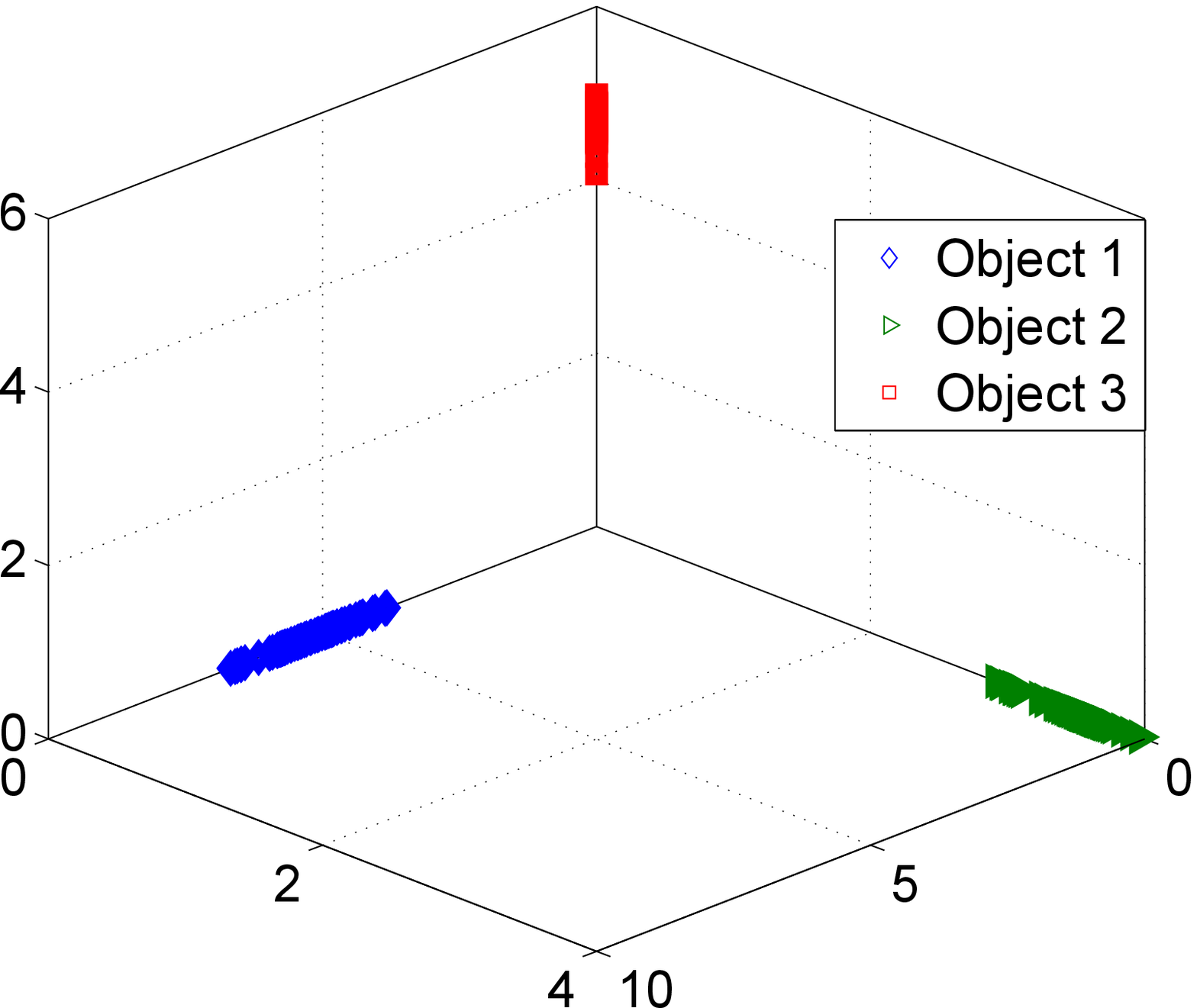}
		\caption{}
		\label{fig:toy_clear}	
	\end{subfigure}
	\begin{subfigure}[b]{0.3\columnwidth}
		\includegraphics[width = \columnwidth]{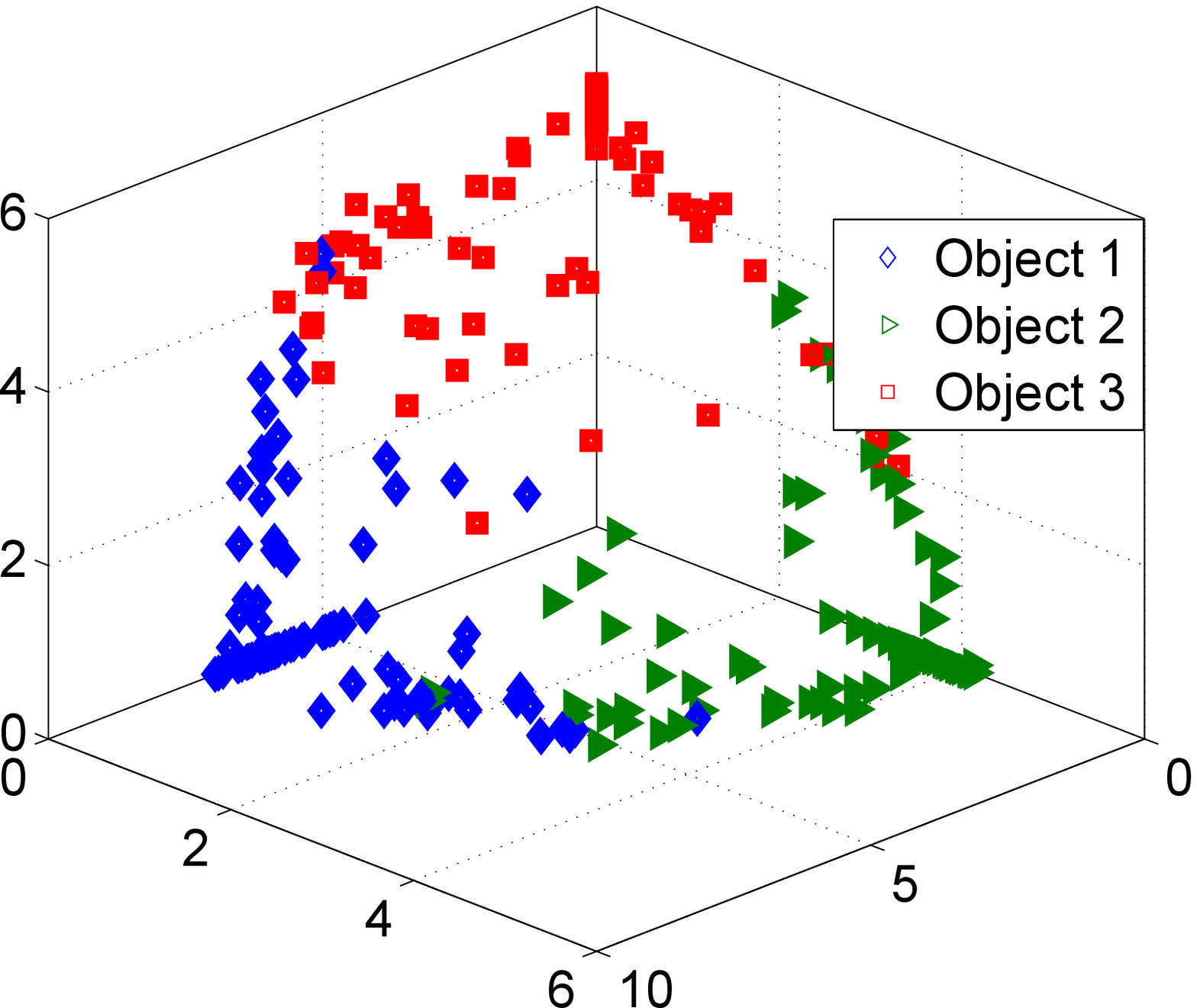}
		\caption{}	
		\label{fig:toy_noisy_bp}
	\end{subfigure}
	\begin{subfigure}[b]{0.3\columnwidth}
		\includegraphics[width = \columnwidth]{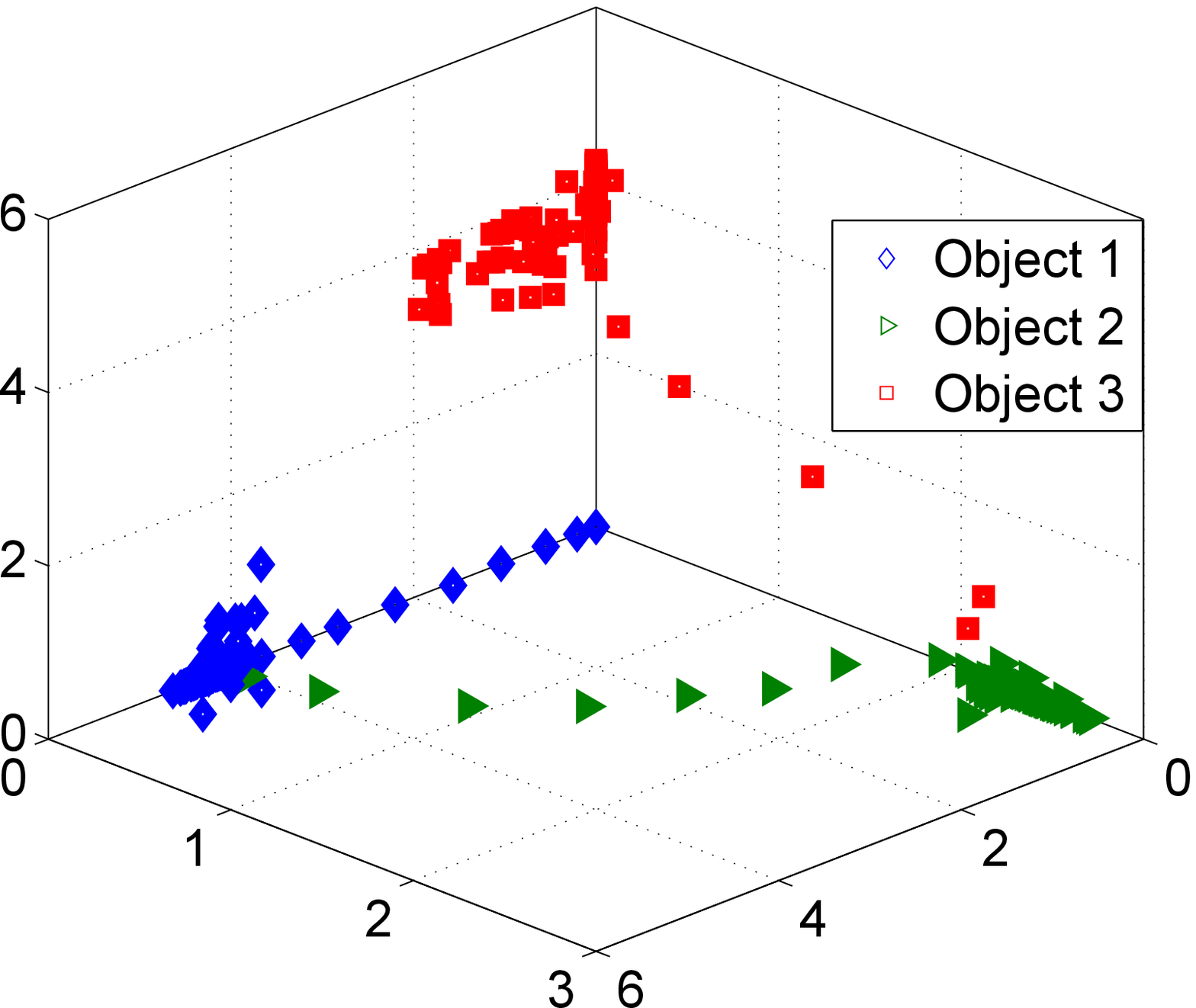}
		\caption{}	
		\label{fig:toy_noisy_td}
	\end{subfigure}
	\caption{Shows the scatter plot of the 3 dimensional causes at the top-layer for (a) clean video with only bottom-up inference, (b) corrupted video with only bottom-up inference and (c) corrupted video with top-down flow along with bottom-up inference. At each point, the shape of the marker indicates the true shape of the object in the frame.} 
\end{figure}

First we consider the clean video and perform inference with only bottom-up inference, i.e., during inference we consider $\mathbf{\hat{u}}_t^{(l)} = 0, \ \forall l \in \{1,2\}$. Figure~\ref{fig:toy_clear} shows the scatter plot of the three dimensional causes at the top layer. Clearly, there are 3 clusters recognizing three different shape in the video sequence. Figure~\ref{fig:toy_noisy_bp} shows the scatter plot when the same procedure is applied on the noisy video. We observe that 3 shapes here can not be clearly distinguished.  Finally, we use top-down information along with the bottom-up inference as described in section \ref{sec:top_down} on the noisy data. We argue that, since the second layer learned class specific information, the top-down information can help the bottom layer units to disambiguate the noisy objects from the true objects.  Figure~\ref{fig:toy_noisy_td} shows the scatter plot for this case. Clearly, with the top-down information, in spite of largely corrupted sequence, the DPCN is able to separate the frames belonging to the three shapes (the trace from one cluster to the other is because of the temporal coherence imposed on the causes at the top layer.). 

\section{Conclusion}
\label{sec:colclusion}
In this paper we proposed the deep predictive coding network, a generative model that empirically alters the priors in a dynamic and context sensitive manner. This model composes to two main components: (a) linear dynamical models with sparse states used for feature extraction, and (b) top-down information to adapt the empirical priors. The  dynamic model captures the temporal dependencies and reduces the instability usually associated with sparse coding \footnote{Please refer to the supplementary material for more details.}, while the task specific information from the top layers helps to resolve ambiguities in the lower-layer improving data representation in the presence of noise. We believe that our approach can be extended with convolutional methods, paving the way for implementation of high-level tasks like object recognition, etc., on large scale videos or images. 

\subsubsection*{Acknowledgments}
This work is supported by the Office of Naval Research (ONR) grant \#N000141010375. We thank Austin J. Brockmeier and Matthew Emigh for their comments and suggestions.
\bibliographystyle{unsrtnat}
\bibliography{DPCN}

\begin{thebibliography}{24}
\providecommand{\natexlab}[1]{#1}
\providecommand{\url}[1]{\texttt{#1}}
\expandafter\ifx\csname urlstyle\endcsname\relax
  \providecommand{\doi}[1]{doi: #1}\else
  \providecommand{\doi}{doi: \begingroup \urlstyle{rm}\Url}\fi

\bibitem[Lowe(1999)]{lowe1999}
David~G. Lowe.
\newblock Object recognition from local scale-invariant features.
\newblock In \emph{Proceedings of the International Conference on Computer
  Vision-Volume 2 - Volume 2}, ICCV '99, pages 1150--, 1999.
\newblock ISBN 0-7695-0164-8.

\bibitem[Dalal and Triggs(2005)]{dalal2005}
Navneet Dalal and Bill Triggs.
\newblock Histograms of oriented gradients for human detection.
\newblock In \emph{Proceedings of the 2005 IEEE Computer Society Conference on
  Computer Vision and Pattern Recognition (CVPR'05) - Volume 1 - Volume 01},
  CVPR '05, pages 886--893, 2005.
\newblock ISBN 0-7695-2372-2.

\bibitem[Olshausen and Field(1996)]{olshausenfield1996}
B.~A. Olshausen and D.~J. Field.
\newblock {Emergence of simple-cell receptive field properties by learning a
  sparse code for natural images.}
\newblock \emph{Nature}, 381\penalty0 (6583):\penalty0 607--609, June 1996.
\newblock ISSN 0028-0836.

\bibitem[Wiskott and Sejnowski(2002)]{wiskott2002slow}
L.~Wiskott and T.J. Sejnowski.
\newblock Slow feature analysis: Unsupervised learning of invariances.
\newblock \emph{Neural computation}, 14\penalty0 (4):\penalty0 715--770, 2002.

\bibitem[Lee et~al.(2009)Lee, Grosse, Ranganath, and Ng]{leeetal2009}
Honglak Lee, Roger Grosse, Rajesh Ranganath, and Andrew~Y. Ng.
\newblock Convolutional deep belief networks for scalable unsupervised learning
  of hierarchical representations.
\newblock In \emph{Proceedings of the 26th Annual International Conference on
  Machine Learning}, ICML '09, pages 609--616, 2009.
\newblock ISBN 978-1-60558-516-1.

\bibitem[Kavukcuoglu et~al.(2010{\natexlab{a}})Kavukcuoglu, Sermanet, Boureau,
  Gregor, Mathieu, and LeCun]{kavukcuoglu2010conv}
K.~Kavukcuoglu, P.~Sermanet, Y.L. Boureau, K.~Gregor, M.~Mathieu, and Y.~LeCun.
\newblock Learning convolutional feature hierarchies for visual recognition.
\newblock \emph{Advances in Neural Information Processing Systems}, pages
  1090--1098, 2010{\natexlab{a}}.

\bibitem[Zeiler et~al.(2010)Zeiler, Krishnan, Taylor, and
  Fergus]{zeiler2010deconvolutional}
M.D. Zeiler, D.~Krishnan, G.W. Taylor, and R.~Fergus.
\newblock Deconvolutional networks.
\newblock In \emph{Computer Vision and Pattern Recognition (CVPR), 2010 IEEE
  Conference on}, pages 2528--2535. IEEE, 2010.

\bibitem[Vincent et~al.(2010)Vincent, Larochelle, Lajoie, Bengio, and
  Manzagol]{vincent2010stacked}
P.~Vincent, H.~Larochelle, I.~Lajoie, Y.~Bengio, and P.A. Manzagol.
\newblock Stacked denoising autoencoders: Learning useful representations in a
  deep network with a local denoising criterion.
\newblock \emph{The Journal of Machine Learning Research}, 11:\penalty0
  3371--3408, 2010.

\bibitem[Rao and Ballard(1997)]{raoballard1997}
Rajesh P.~N. Rao and Dana~H. Ballard.
\newblock Dynamic model of visual recognition predicts neural response
  properties in the visual cortex.
\newblock \emph{Neural Computation}, 9:\penalty0 721--763, 1997.

\bibitem[Friston(2008)]{friston2008}
Karl Friston.
\newblock Hierarchical models in the brain.
\newblock \emph{PLoS Comput Biol}, 4\penalty0 (11):\penalty0 e1000211, 11 2008.

\bibitem[Hinton et~al.()Hinton, Osindero, and Teh]{hinton2006}
Geoffrey~E. Hinton, Simon Osindero, and Yee-Whye Teh.
\newblock {A Fast Learning Algorithm for Deep Belief Nets}.
\newblock \emph{Neural Comp.}, \penalty0 (7):\penalty0 1527--1554, July .

\bibitem[Bengio et~al.(2007)Bengio, Lamblin, Popovici, and
  Larochelle]{bengio2006}
Yoshua Bengio, Pascal Lamblin, Dan Popovici, and Hugo Larochelle.
\newblock {Greedy layer-wise training of deep networks}.
\newblock In \emph{In NIPS}, 2007.

\bibitem[Kavukcuoglu et~al.(2010{\natexlab{b}})Kavukcuoglu, Ranzato, and
  LeCun]{kavukcuoglu2010fast}
Koray Kavukcuoglu, Marc'Aurelio Ranzato, and Yann LeCun.
\newblock Fast inference in sparse coding algorithms with applications to
  object recognition.
\newblock \emph{CoRR}, abs/1010.3467, 2010{\natexlab{b}}.

\bibitem[Karklin and Lewicki(2005)]{karklin2005}
Yan Karklin and Michael~S. Lewicki.
\newblock A hierarchical bayesian model for learning nonlinear statistical
  regularities in nonstationary natural signals.
\newblock \emph{Neural Computation}, 17:\penalty0 397--423, 2005.

\bibitem[Angelosante et~al.(2009)Angelosante, Giannakis, and
  Grossi]{angelo2009}
D.~Angelosante, G.B. Giannakis, and E.~Grossi.
\newblock Compressed sensing of time-varying signals.
\newblock In \emph{Digital Signal Processing, 2009 16th International
  Conference on}, pages 1 --8, july 2009.

\bibitem[Charles et~al.(2011)Charles, Asif, Romberg, and Rozell]{charles2011}
A.~Charles, M.S. Asif, J.~Romberg, and C.~Rozell.
\newblock Sparsity penalties in dynamical system estimation.
\newblock In \emph{Information Sciences and Systems (CISS), 2011 45th Annual
  Conference on}, pages 1 --6, march 2011.

\bibitem[Sejdinovic et~al.(2010)Sejdinovic, Andrieu, and
  Piechocki]{sejdinovic2010}
D.~Sejdinovic, C.~Andrieu, and R.~Piechocki.
\newblock Bayesian sequential compressed sensing in sparse dynamical systems.
\newblock In \emph{Communication, Control, and Computing (Allerton), 2010 48th
  Annual Allerton Conference on}, pages 1730 --1736, 29 2010-oct. 1 2010.
\newblock \doi{10.1109/ALLERTON.2010.5707125}.

\bibitem[Chen et~al.(2012)Chen, Lin, Kim, Carbonell, and
  Xing]{chen2012smoothing}
X.~Chen, Q.~Lin, S.~Kim, J.G. Carbonell, and E.P. Xing.
\newblock Smoothing proximal gradient method for general structured sparse
  regression.
\newblock \emph{The Annals of Applied Statistics}, 6\penalty0 (2):\penalty0
  719--752, 2012.

\bibitem[Beck and Teboulle()]{fista2009}
Amir Beck and Marc Teboulle.
\newblock {A Fast Iterative Shrinkage-Thresholding Algorithm for Linear Inverse
  Problems}.
\newblock \emph{SIAM Journal on Imaging Sciences}, \penalty0 (1):\penalty0
  183--202, March .
\newblock ISSN 19364954.
\newblock \doi{10.1137/080716542}.

\bibitem[Nesterov(2005)]{nesterov2005smooth}
Y.~Nesterov.
\newblock Smooth minimization of non-smooth functions.
\newblock \emph{Mathematical Programming}, 103\penalty0 (1):\penalty0 127--152,
  2005.

\bibitem[Gregor and LeCun(2011)]{gregor2011}
Karol Gregor and Yann LeCun.
\newblock {Efficient Learning of Sparse Invariant Representations}.
\newblock \emph{CoRR}, abs/1105.5307, 2011.

\bibitem[Nelson(2000)]{nelson2000}
Alex Nelson.
\newblock \emph{Nonlinear estimation and modeling of noisy time-series by dual
  Kalman filtering methods}.
\newblock PhD thesis, 2000.

\bibitem[LeCun et~al.(1989)LeCun, Boser, Denker, Henderson, Howard, Hubbard,
  and Jackel]{lecun1989}
Y.~LeCun, B.~Boser, J.~S. Denker, D.~Henderson, R.~E. Howard, W.~Hubbard, and
  L.~D. Jackel.
\newblock {Backpropagation applied to handwritten zip code recognition}.
\newblock \emph{Neural Comput.}, 1\penalty0 (4):\penalty0 541--551, December
  1989.
\newblock ISSN 0899-7667.
\newblock \doi{10.1162/neco.1989.1.4.541}.
\newblock URL \url{http://dx.doi.org/10.1162/neco.1989.1.4.541}.

\bibitem[Zou and Hastie(2005)]{zou2005regularization}
H.~Zou and T.~Hastie.
\newblock Regularization and variable selection via the elastic net.
\newblock \emph{Journal of the Royal Statistical Society: Series B (Statistical
  Methodology)}, 67\penalty0 (2):\penalty0 301--320, 2005.

\end{thebibliography}

\vfill\pagebreak
\setcounter{section}{0}
\renewcommand\thesection{\Alph{section}}

\section{Supplementary material for Deep Predictive Coding Networks}
\subsection{From section~\ref{sec:infer}, computing $\boldsymbol{\alpha}^{*}$}
The optimal solution of $\boldsymbol{\alpha}$ in  (\ref{eq:nestrov_approx}) is given by
\begin{align}
	\boldsymbol{\alpha}^{*} = & \argmax\limits_{\|\boldsymbol{\alpha}\|_{\infty} \leq 1} [\boldsymbol{\alpha}^{T}\mathbf{e}_t - \frac{\mu}{2} \|\boldsymbol{\alpha}\|^2] \nonumber \\
	 = & \argmin\limits_{\|\boldsymbol{\alpha}\|_{\infty} \leq 1} \Big\|\boldsymbol{\alpha} - \frac{\mathbf{e}_t}{\mu}\Big\|^2 \nonumber \\
	 = & S\Big(\frac{\mathbf{e}_t}{\mu}\Big)
	 \label{eq:alpha_update}
\end{align}
where $S(.)$ is a function projecting $\Big(\frac{\mathbf{e}_t}{\mu}\Big)$ onto an $\ell_{\infty}$-ball. This is of the form:
\begin{align*}
	S(x) = &
	\begin{cases}
	x, & -1 \leq x \leq 1 \\
	1, & x > 1 \\
	-1, & x < -1
	\end{cases}
\end{align*}

\subsection{Algorithm for joint inference of the states and the causes.}

\begin{algorithm}[h]
\caption{Updating $\mathbf{x}_t$,$\mathbf{u}_t$ simultaneously using FISTA-like procedure \citep{fista2009}.}
\label{algo:fista}
\begin{algorithmic}[1]
\Require Take $L_{0,n}^x > 0 \;\forall n \in \{1,2,...,N\}$, $L_0^u > 0$ and some $\eta > 1$.
\State Initialize $\mathbf{x}_{0,n} \in \mathbb{R}^{K}\; \forall n \in \{1,2,...,N\}$, $\mathbf{u}_0 \in \mathbb{R}^{D}$ and set $\xi_1 = \mathbf{u}_0$, $\mathbf{z}_{1,n} = \mathbf{x}_{0,n}$. 
\State Set step-size parameters: $\tau_{1} = 1$.
	\While{no convergence}
		\State Update $$\boldsymbol{\gamma} = \gamma_0(1 + \exp(-[\mathbf{B}\mathbf{u}_i])/2$$.
		\For{$n \in \{1,2,...,N\}$}
		\State \textbf{Line search:} Find the best step size $L_{k,n}^x$.
		\State Compute $\boldsymbol{\alpha}^{*}$ from (\ref{eq:alpha_update})
		\State Update $\mathbf{x}_{i,n}$ using the gradient from (\ref{eq:state_grad}) with a soft-thresholding function.
		\State Update internal variables $z_{i+1}$ with step size parameter $\tau_{i}$ as in \citep{fista2009}.
		\EndFor
		\State Compute $\sum_{n=1}^N |\mathbf{x}_{i,n}|$
		\State \textbf{Line search:} Find the best step size $L_{k}^u$.
		\State Update $\mathbf{u}_{i,n}$ using the gradient of (\ref{eq:causes_smooth}) with a soft-thresholding function.
		\State Update internal variables $\xi_{i+1}$ with step size parameter $\tau_{i}$ as in \citep{fista2009}.
		\State Update $$\tau_{i+1} = \Big(1 + \sqrt{(4\tau_{i}^2+1)}\Big)/2$$.
		\State Check for convergence.
		\State $i = i + 1$
	\EndWhile  
\\ \Return $\mathbf{x}_{i,n}\ \forall n \in \{1,2,...,N\}$ and $\mathbf{u}_i$
\end{algorithmic}
\end{algorithm}

\vfill\pagebreak

\subsection{Inferring sparse states with known parameters}
\begin{figure}[h]
\centering
	\includegraphics[width = 0.8\columnwidth]{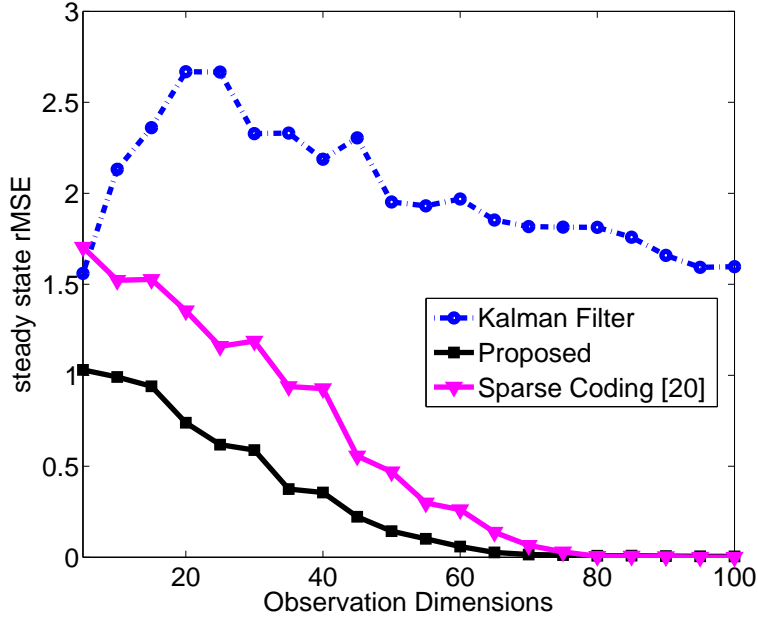}
	\caption{Shows the performance of the inference algorithm with fixed parameters when compared with sparse coding and Kalman filtering. For this we first simulate a state sequence with only 20 non-zero elements in a 500-dimensional state vector evolving with a permutation matrix, which is different for every time instant, followed by a scaling matrix to generate a sequence of observations. We consider that both the permutation and the scaling matrices are known \emph{apriori}. The observation noise is Gaussian zero mean and variance $\sigma^2 = 0.01$. We consider sparse state-transition noise, which is simulated by choosing a subset of active elements in the state vector (number of elements is chosen randomly via a Poisson distribution with mean 2) and switching each of them with a randomly chosen element (with uniform probability over the state vector). This resemble a sparse innovation in the states. We use these generated observation sequences as inputs and use the \emph{apriori} know parameters to infer the states from the dynamic model. 
Figure~\ref{fig:state_est} shows the results obtained, where we compare the inferred states from different methods with the true states in terms of relative mean squared error (rMSE) (defined as $\|\mathbf{{x}}_t^{est} - \mathbf{x}_{t}^{true}\| / \|\mathbf{x}_t^{true}\|$). The steady state error (rMSE) after 50 time instances is plotted versus with the dimensionality of the observation sequence. Each point is obtained after averaging over 50 runs. We observe that our model is able to converge to the true solution even for low dimensional observation, when other methods like sparse coding fail. We argue that the temporal dependencies considered in our model is able to drive the solution to the right attractor basin, insulating it from instabilities typically associated with sparse coding \citep{zou2005regularization}. }
	\label{fig:state_est}
\end{figure}

\vfill\pagebreak

\subsection{Visualizing first layer of the learned model}
\begin{figure}[h]
\centering
	\begin{subfigure}[b]{0.4\textwidth}
	\centering
		\includegraphics[width = 5cm, height = 5cm]{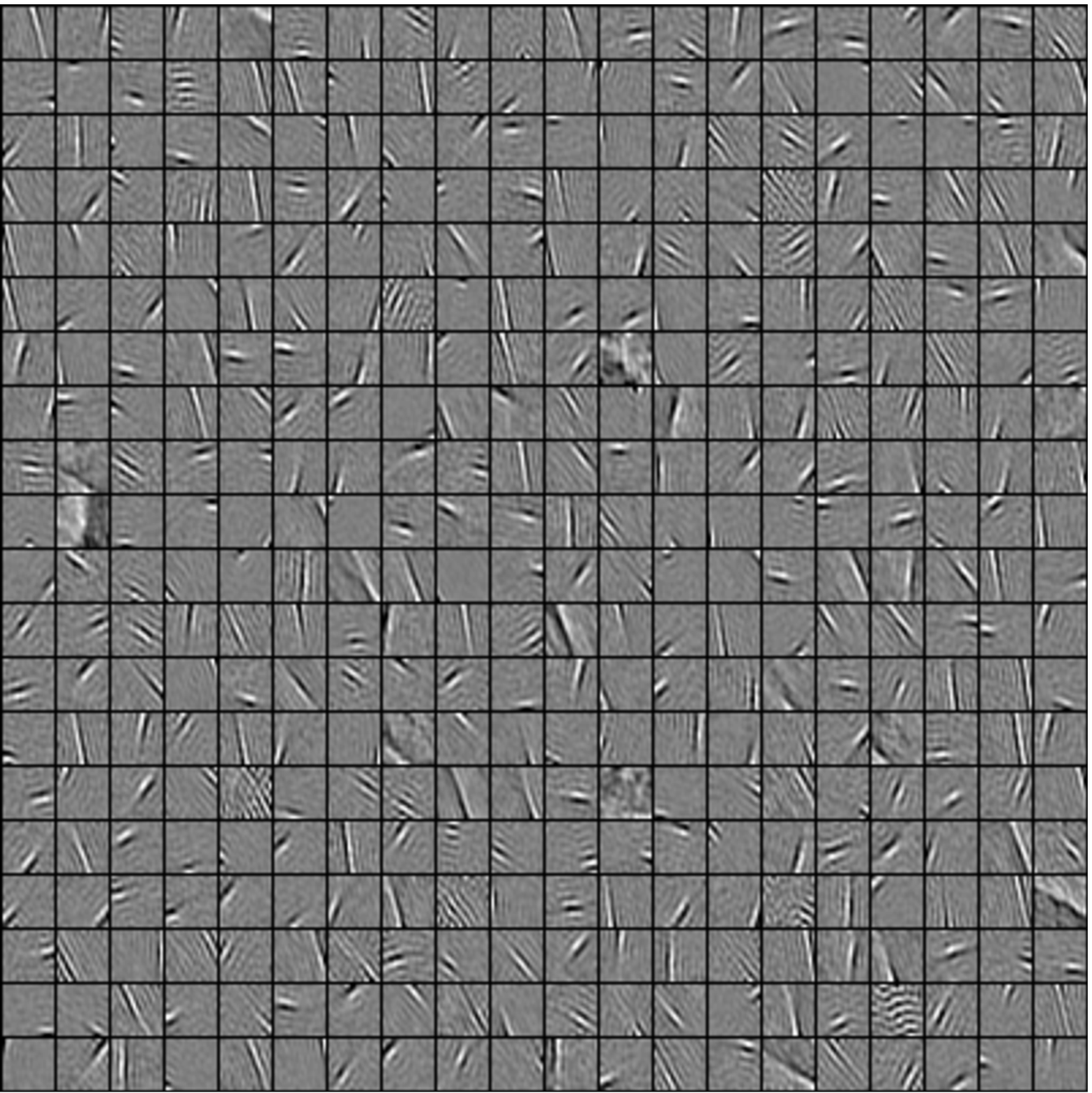}	
		\caption{Observation matrix (Bases)}		
		\label{fig:basis_states}
	\end{subfigure}
	\begin{subfigure}[b]{0.55\textwidth}
		\centering
		\includegraphics[width = 7cm, height = 5cm]{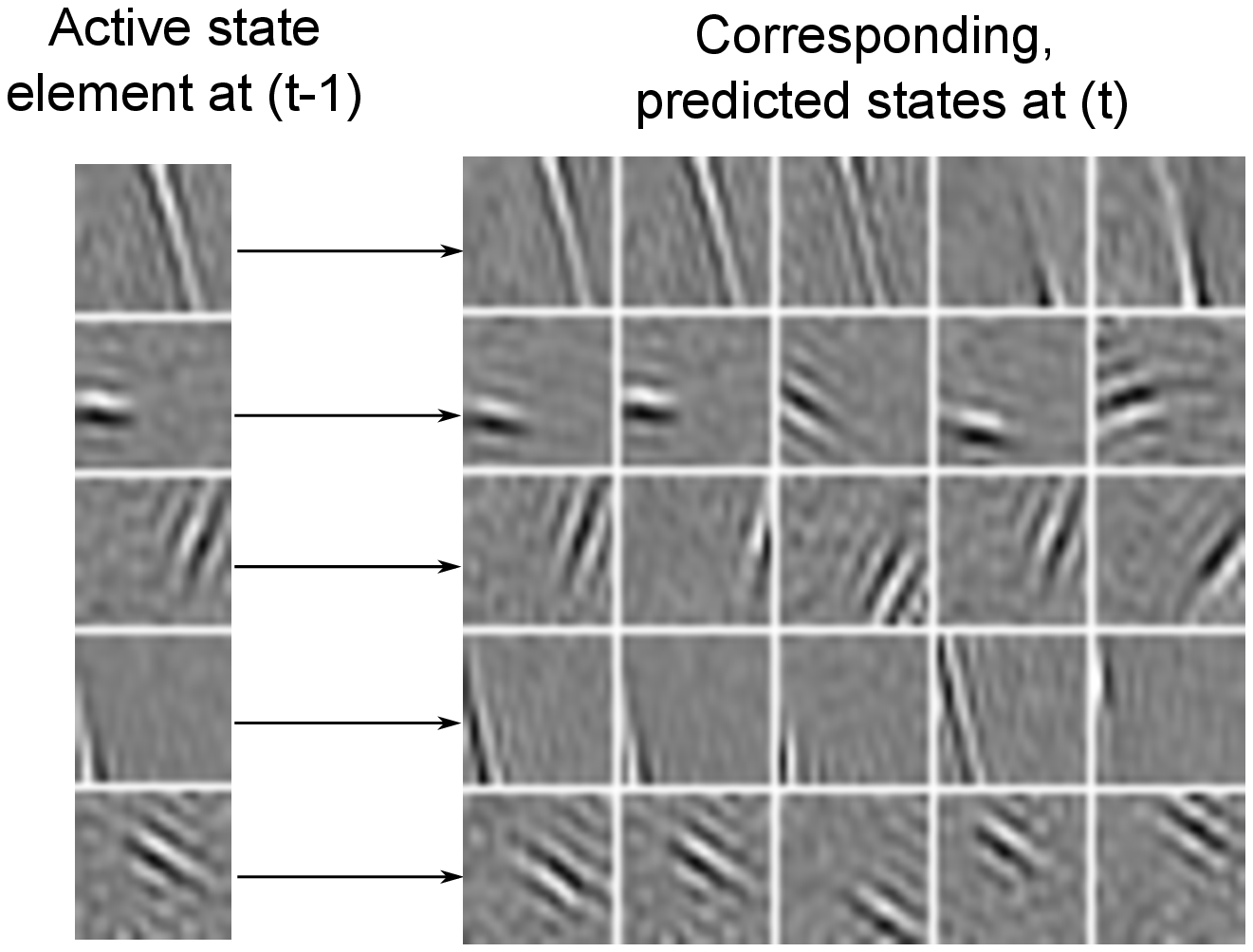}
		\caption{State-transition matrix}
		\label{fig:state_prediction}
	\end{subfigure}
	\caption{Visualization of the parameters. $\mathbf{C}$ and $\mathbf{A}$, of the model described in section \ref{sec:results_recept}. (A) Shows the learned observation matrix $\mathbf{C}$. Each square block indicates a column of the matrix, reshaped as $\sqrt{p} \times \sqrt{p}$ pixel block. (B) Shows the state transition matrix $\mathbf{A}$ using its connections strength with the observation matrix $\mathbf{C}$. On the left are the basis corresponding to the single active element in the state at time $(t-1)$ and on the right are the basis corresponding to the five most ``active" elements in the predicted state (ordered in decreasing order of the magnitude).}
\end{figure}   
  
\begin{figure}[hb]
\centering
	\begin{subfigure}[b]{0.3\textwidth}
		\centering
		\includegraphics[width = 2.5cm, height = 4.5cm]{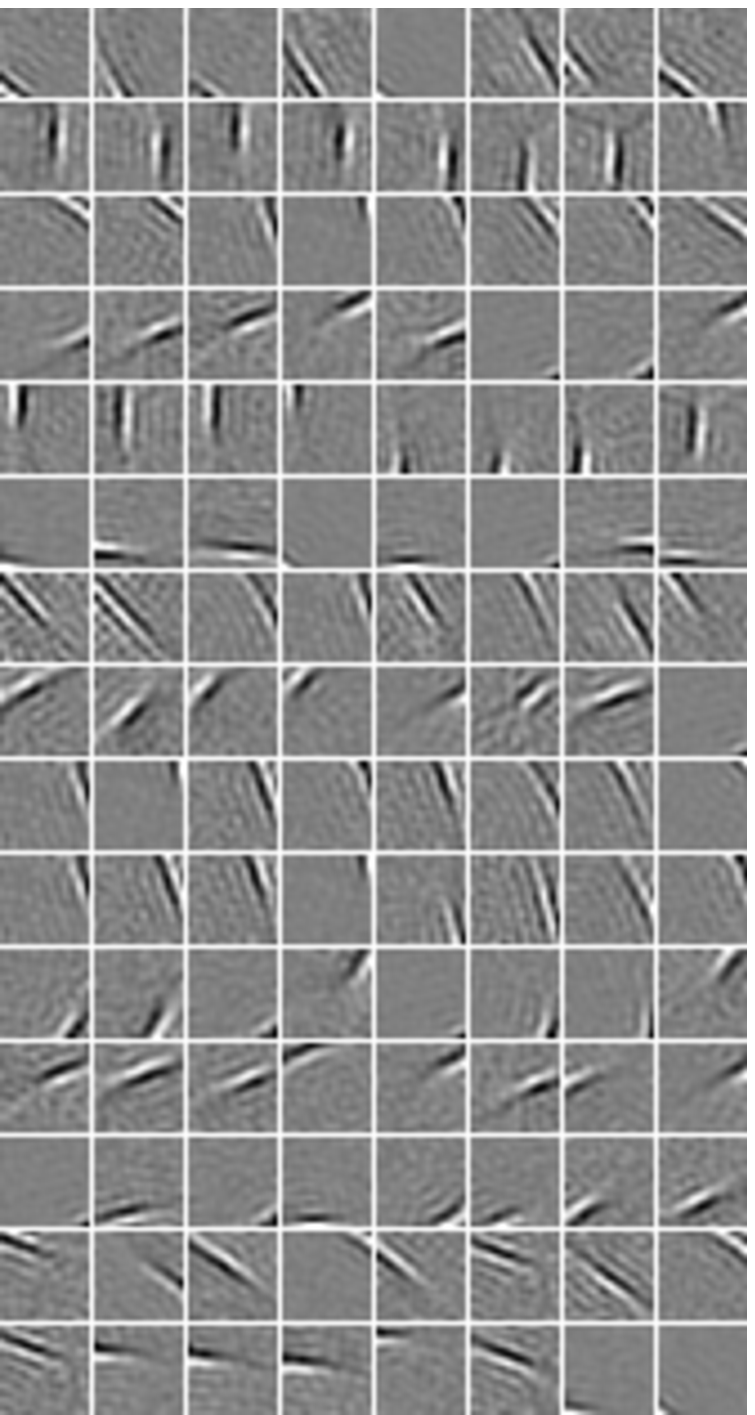}
		\caption{Connections}
	\end{subfigure}
	\hspace*{-7mm}
	\begin{subfigure}[b]{0.35\textwidth}
		\centering
		\includegraphics[width = 4.7cm, height = 4.5cm]{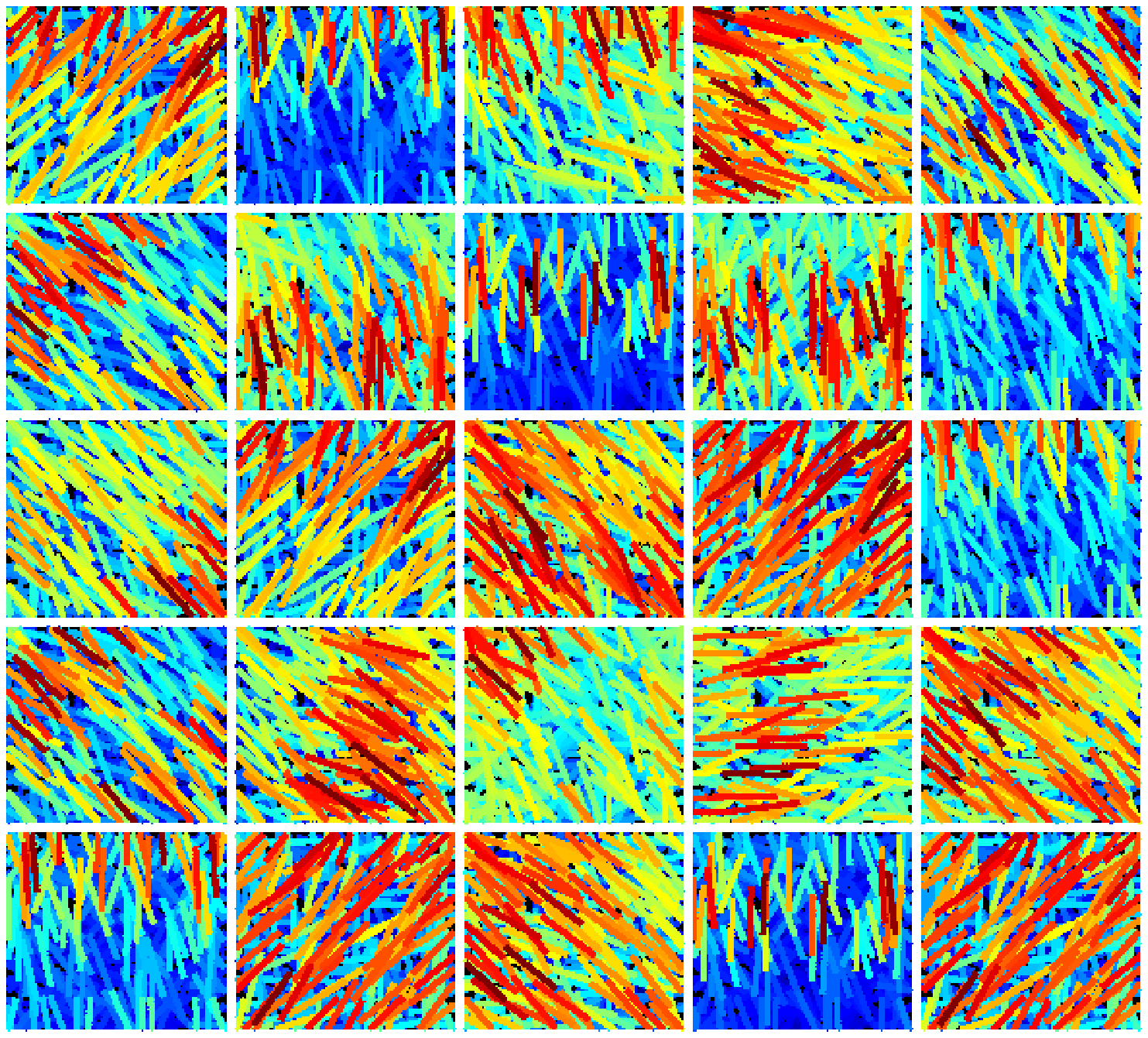}
		\caption{Centers and Orientations}
	\end{subfigure}
	\begin{subfigure}[b]{0.35\textwidth}
	\centering
		\includegraphics[width = 4.7cm, height = 4.5cm]{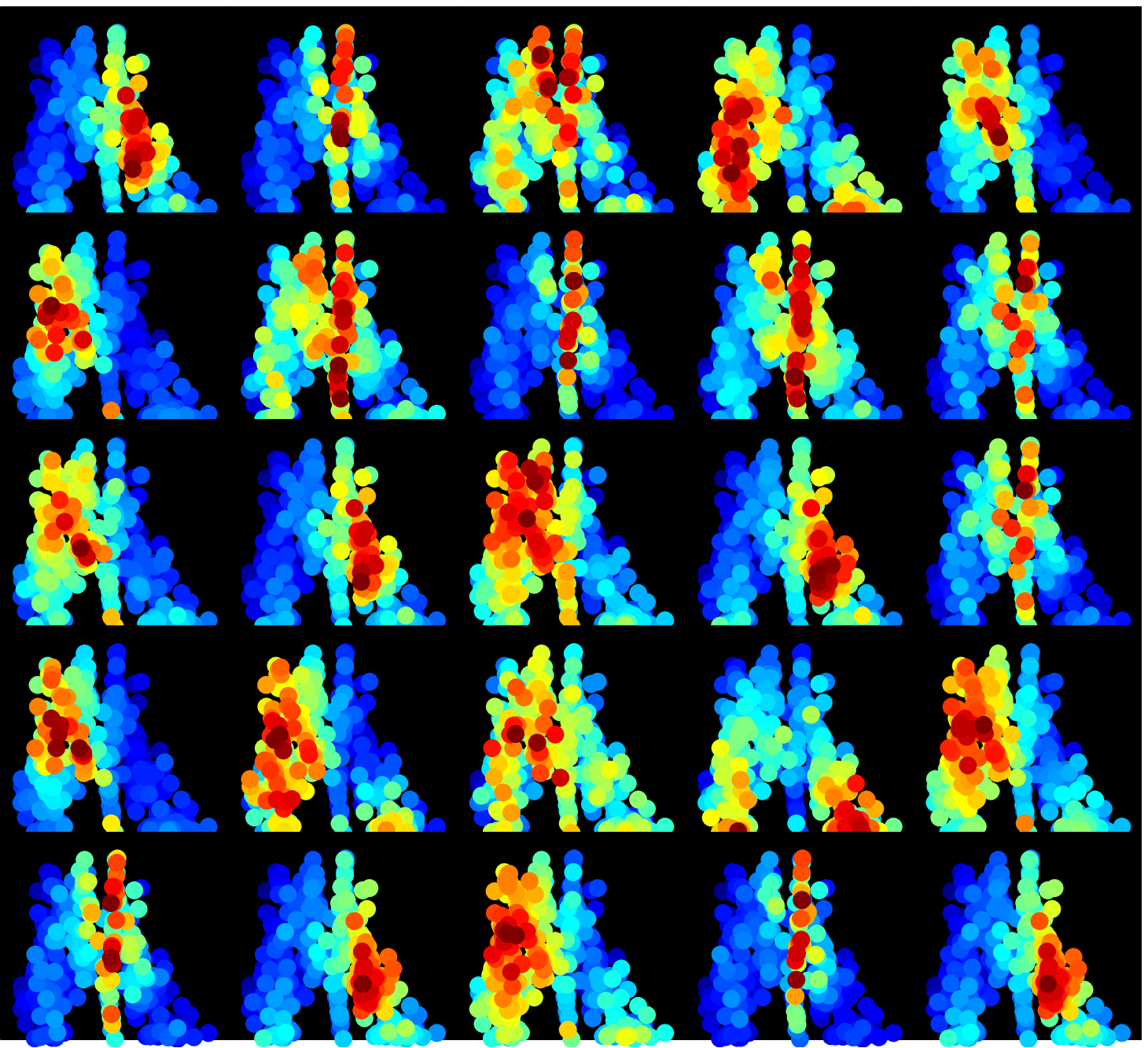}
		\caption{Orientations and Frequencies}
	\end{subfigure}
	\caption[Connections between the invariant units and the basis functions. (A) Shows the connections between the basis and columns of $\mathbf{B}$.  (B) Shows spatially localized grouping of the invariant units. (C) Similarly, we show the corresponding orientation and spatial frequency selectivity of the invariant units. ]{Connections between the invariant units and the basis functions. (A) Shows the connections between the basis and columns of $\mathbf{B}$. Each row indicates an invariant unit. Here the set of basis that a strongly correlated to an invariant unit are shown, arranged in the decreasing order of the magnitude. (B) Shows spatially localized grouping of the invariant units. Firstly, we fit a Gabor function to each of the basis functions. Each subplot here is then obtained by plotting a line indicating the center and the orientation of the Gabor function. The colors indicate the connections strength with an invariant unit; red indicating stronger connections and blue indicate almost zero strength. We randomly select a subset of 25 invariant units here. We observe that the invariant unit group the basis that are local in spatial centers and orientations. (C) Similarly, we show the corresponding orientation and spatial frequency selectivity of the invariant units.  Here each plot indicates the orientation and frequency of each Gabor function color coded according to the connection strengths with the invariant units. Each subplot is a half-polar plot with the orientation plotted along the angle ranging from $0$ to $\pi$ and the distance from the center indicating the frequency. Again, we observe that the invariant units group the basis that have similar orientation.}
	\end{figure}
\end{document}